\documentclass{article}

\usepackage[numbers]{natbib}
\usepackage[preprint]{neurips_2022}

\usepackage[dvipsnames]{xcolor}         
\definecolor{linkColor}{rgb}{0.18,0.39,0.62}
\usepackage[utf8]{inputenc} 
\usepackage[T1]{fontenc}    
\usepackage[colorlinks=true,linkcolor=linkColor,citecolor=linkColor,filecolor=linkColor,urlcolor=linkColor]{hyperref}       
\usepackage{url}            
\usepackage{booktabs}       
\usepackage{amsfonts}       
\usepackage{nicefrac}       
\usepackage{microtype}      

\usepackage{graphicx}
\usepackage{arydshln}
\usepackage{booktabs}
\usepackage{multirow}
\usepackage{caption}
\usepackage{subcaption}
\usepackage{makecell}
\usepackage{csquotes}
\usepackage{epigraph}
\usepackage{amsmath}
\usepackage{xcolor}  
\usepackage{cleveref}
\usepackage{pifont}
\usepackage{listings}

\RequirePackage{algorithm}
\RequirePackage{algorithmic}

\usepackage{multirow}
\usepackage{amsmath}
\usepackage{capt-of}
\usepackage{tabularx}
\usepackage{epsfig}
\usepackage{amssymb}
\usepackage{amsfonts}
\usepackage{booktabs}
\usepackage{scalerel}
\usepackage[inline]{enumitem}
\usepackage{listings}
\usepackage{varwidth}
\usepackage[export]{adjustbox}
\usepackage{tikz}
\usetikzlibrary{tikzmark}

\usepackage{stmaryrd}
\usepackage{bbm}
\usepackage{wrapfig}
\usepackage{pifont}

\definecolor{deepblue}{rgb}{0,0,0.5}
\definecolor{officeblue}{RGB}{0,102,204}
\definecolor{deepred}{rgb}{0.6,0,0}
\definecolor{deepgreen}{rgb}{0,0.5,0}
\definecolor{mybrickred}{RGB}{182,50,28}

\definecolor{fillcolor}{RGB}{216,217,252}

\usepackage{etoolbox}
\usepackage{framed}

\newif\ifxetexorluatex
\ifxetex
  \xetexorluatextrue
\else
  \ifluatex
    \xetexorluatextrue
  \else
    \xetexorluatexfalse
  \fi
\fi
\newcommand*\quotesize{60} 
\newcommand*{\openquote}
   {\tikz[remember picture,overlay,xshift=-4ex,yshift=-2.5ex]
   \node (OQ) {\fontsize{\quotesize}{\quotesize}\selectfont``};\kern0pt}

\newcommand*{\closequote}[1]
  {\tikz[remember picture,overlay,xshift=4ex,yshift={#1}]
   \node (CQ) {\fontsize{\quotesize}{\quotesize}\selectfont''};}

\colorlet{shadecolor}{white}

\newcommand*\shadedauthorformat{\emph} 

\newcommand*\authoralign[1]{%
  \if#1l
    \def\authorfill{}\def\quotefill{\hfill}
  \else
    \if#1r
      \def\authorfill{\hfill}\def\quotefill{}
    \else
      \if#1c
        \gdef\authorfill{\hfill}\def\quotefill{\hfill}
      \else\typeout{Invalid option}
      \fi
    \fi
  \fi}
\newenvironment{shadequote}[2][l]%
{\authoralign{#1}
\ifblank{#2}
   {\def\shadequoteauthor{}\def\yshift{-2ex}\def\quotefill{\hfill}}
   {\def\shadequoteauthor{\par\authorfill\shadedauthorformat{#2}}\def\yshift{2ex}}
\begin{snugshade}\begin{quote}\openquote}
{\shadequoteauthor\quotefill\closequote{\yshift}\end{quote}\end{snugshade}}

\usepackage{amsmath,amsfonts,bm}

\def\eqref#1{equation~\ref{#1}}

\def\1{\bm{1}}

\DeclareMathAlphabet{\mathsfit}{\encodingdefault}{\sfdefault}{m}{sl}
\SetMathAlphabet{\mathsfit}{bold}{\encodingdefault}{\sfdefault}{bx}{n}

\newcommand{\cmark}{\ding{51}}%
\newcommand{\xmark}{\ding{55}}%

\newcommand\our{BitNet}

\title{\our{}: Scaling \colorbox{gray!30}{1-bit} Transformers for \\ Large Language Models}

\author{
Hongyu Wang\thanks{~Equal contribution. $\diamond$ Corresponding author.}$~~^{\dag\ddag}$~~~~Shuming Ma\footnotemark[1]$~~^{\dag}$~~~~Li Dong$^{\dag}$~~~Shaohan Huang$^{\dag}$ \\
\bf Huaijie Wang$^{\mathsection}$~~~Lingxiao Ma$^{\dag}$~~~Fan Yang$^{\dag}$~~~Ruiping Wang$^{\ddag}$~~~Yi Wu$^{\mathsection}$~~~Furu Wei$^{\dag}$$^{\diamond}$ \\
$^\dag$ Microsoft Research ~~~
$^\ddag$ University of Chinese Academy of Sciences ~~ 
$^\mathsection$ Tsinghua University \\
{\href{https://aka.ms/GeneralAI}{https://aka.ms/GeneralAI}}
\vspace{-0.4cm}
\\}

\begin{document}

\maketitle

\begin{abstract}
The increasing size of large language models has posed challenges for deployment and raised concerns about environmental impact due to high energy consumption.
In this work, we introduce \our{}, a scalable and stable 1-bit Transformer architecture designed for large language models.
Specifically, we introduce \texttt{BitLinear} as a drop-in replacement of the \texttt{nn.Linear} layer in order to train 1-bit weights from scratch.
Experimental results on language modeling show that \our{} achieves competitive performance while substantially reducing memory footprint and energy consumption, compared to state-of-the-art 8-bit quantization methods and FP16 Transformer baselines.
Furthermore, \our{} exhibits a scaling law akin to full-precision Transformers, suggesting its potential for effective scaling to even larger language models while maintaining efficiency and performance benefits.
\end{abstract}

\vfill

\begin{figure}[ht]
\centering
\begin{subfigure}{.4\textwidth}
\centering
\includegraphics[width=\linewidth]{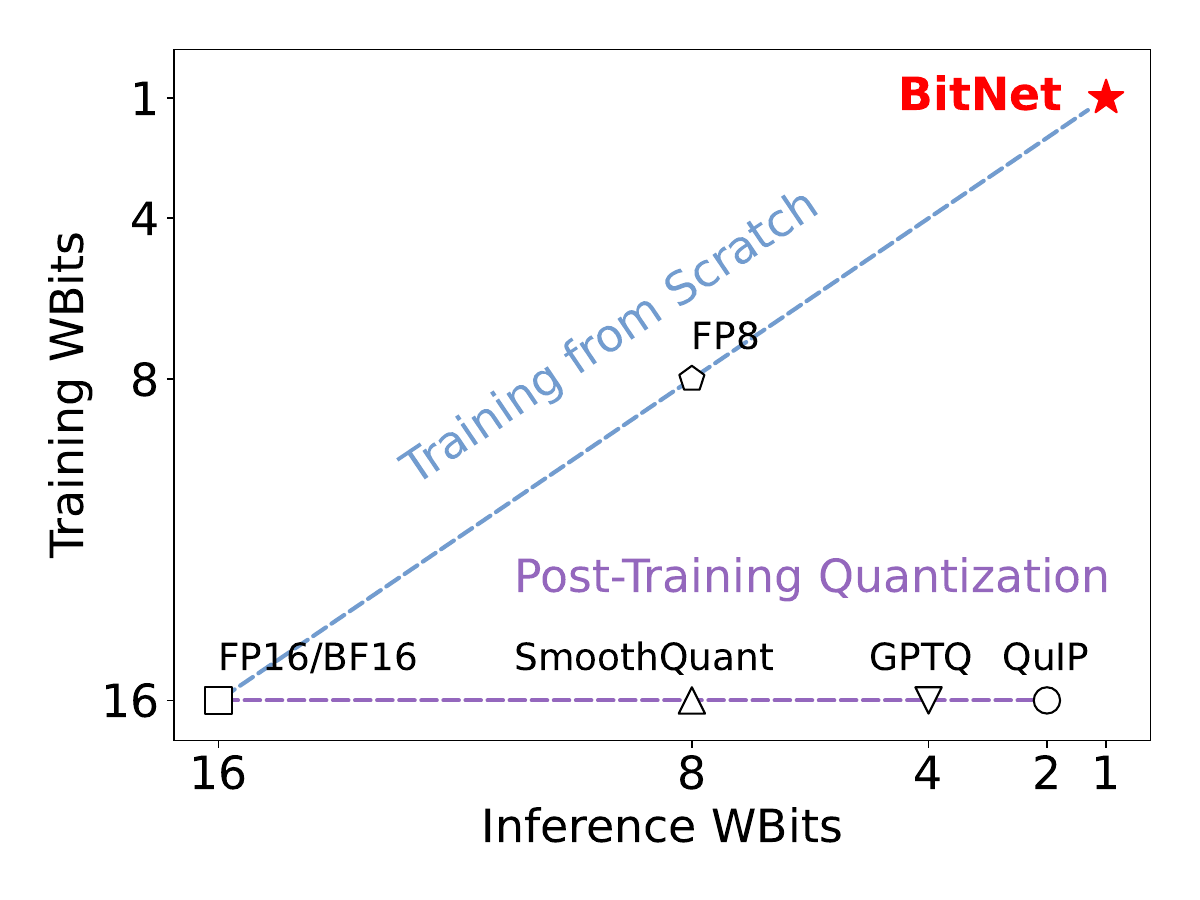}
\end{subfigure}
\begin{subfigure}{.52\textwidth}
\centering
\includegraphics[width=\linewidth]{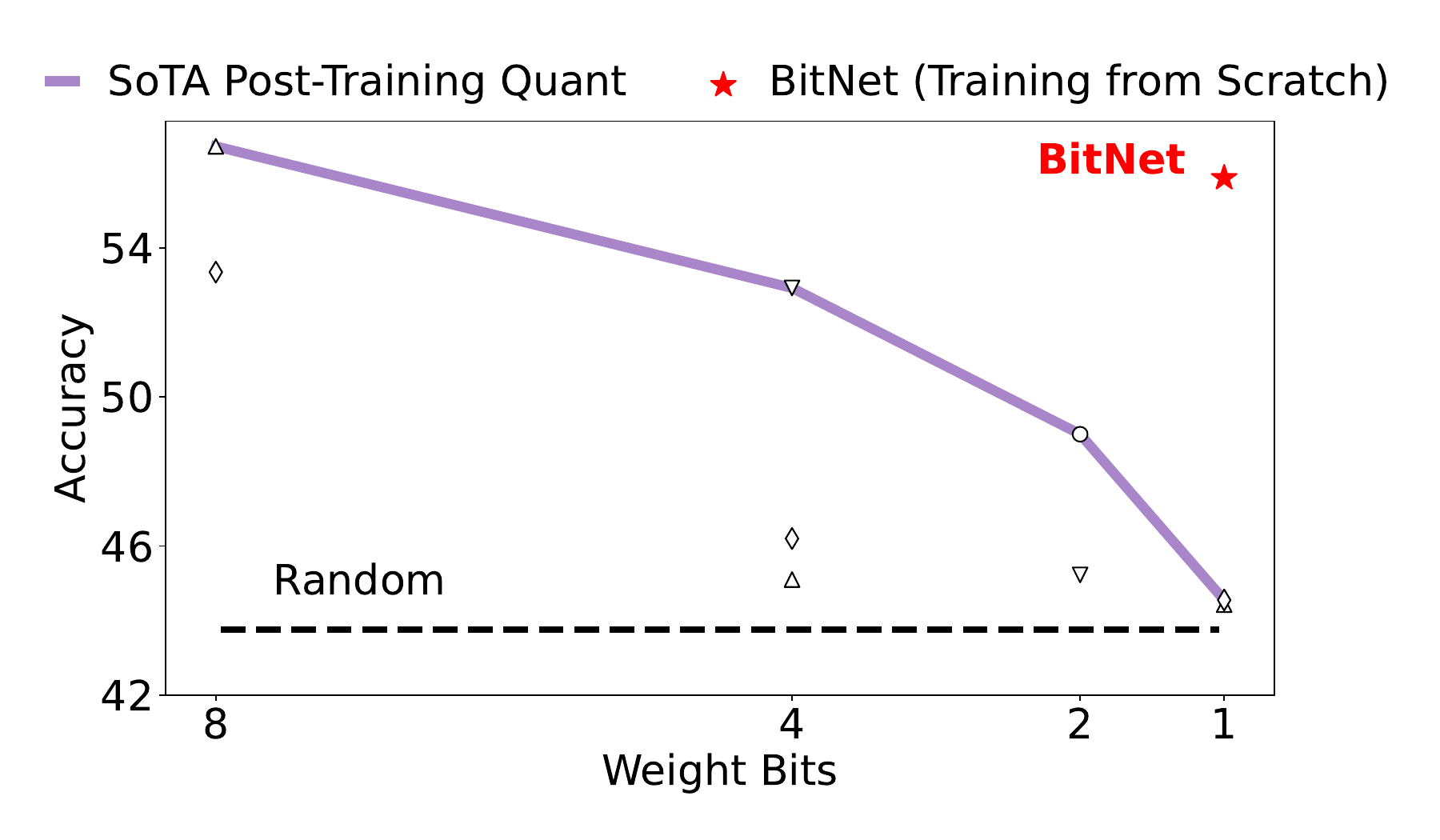}
\end{subfigure}
\begin{subfigure}{.59\textwidth}
\centering
\includegraphics[width=\linewidth]{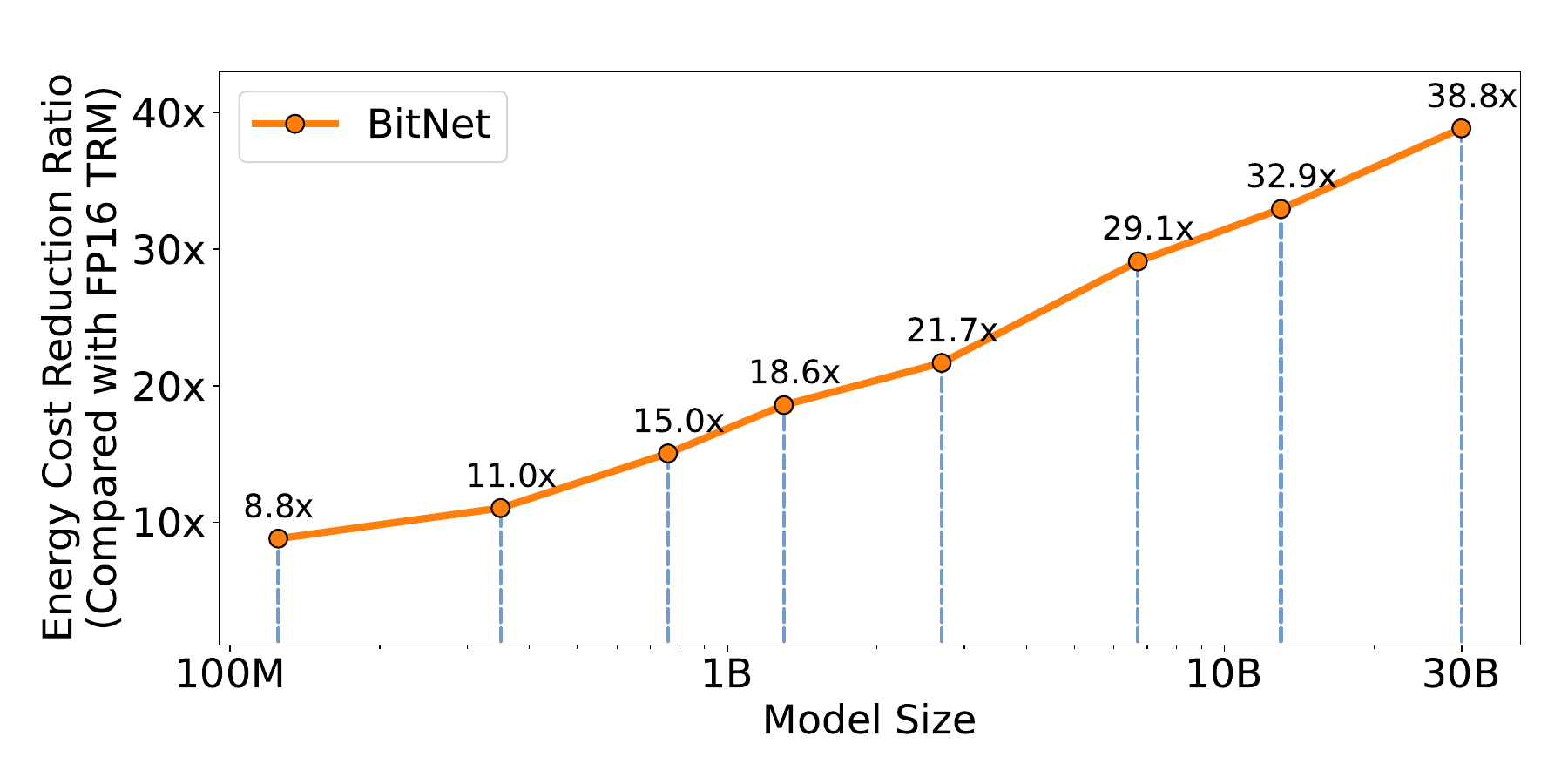}
\end{subfigure}
\begin{subfigure}{.4\textwidth}
\centering
\includegraphics[width=\linewidth]{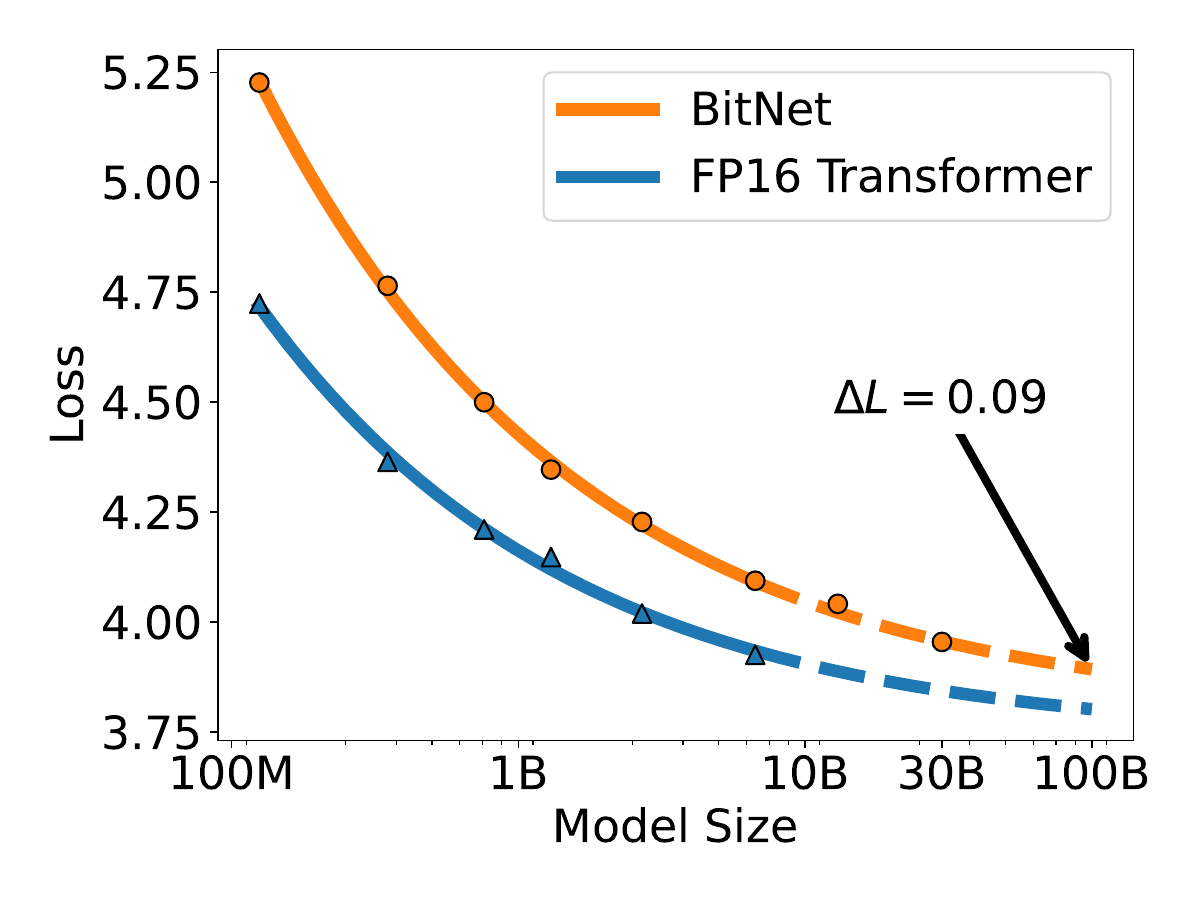}
\end{subfigure}
\caption{\our{} trains 1-bit Transformers from scratch, obtaining competitive results in an energy-efficient way.
\our{} significantly outperforms state-of-the-art quantization methods.
As the model size scales up, the cost savings become more significant while achieving competitive performance with the models trained with FP16.}
\end{figure}

\newpage

\begin{shadequote}[r]{\small William Henry Gates III}
{\small I don't think there's anything unique about human intelligence. All the neurons in the brain that make up perceptions and emotions operate in a binary fashion.}
\end{shadequote}

\section{Introduction}

The rapid growth of large language models~\cite{gpt3, gpt4, palm, palm2, llama, llama2} has led to significant improvements in various tasks.
However, it is expensive to host large language models due to the high inference costs and energy consumption.
As the size of these models grows, the memory bandwidth required for accessing and processing the model parameters becomes a major bottleneck, limiting the overall inference performance.
Moreover, when deploying these models on distributed systems or multi-device platforms, the inter-device communication overhead can significantly impact the inference latency and energy consumption.
Model quantization~\cite{gptq, quip, smoothquant} has emerged as a promising solution, as it can significantly reduce the memory footprint and computational cost of large-scale models while maintaining competitive performance.

Most existing quantization approaches for large language models are post-training. They are simple and easy to apply since it does not require any changes to the training pipeline or retraining the model. However, it will result in a more significant loss of accuracy especially when the precision goes lower, because the model is not optimized for the quantized representation during training.

Another strand of quantizing deep neural networks is quantization-aware training. Compared to post-training, it typically results in better accuracy, as the model is trained to account for the reduced precision from the beginning. Moreover, it allows the model to continue-train or do fine-tuning, which is essential for large language models. The challenge of quantization-aware training mainly lies in optimization, i.e., the model becomes more difficult to converge as the precision goes lower. Besides, it is unknown whether quantization-aware training follows the scaling law of neural language models.

In this work, we focus on binarization (i.e., 1-bit), which is the extreme case of quantization, applied to large language models. Previous studies on binarized neural networks~\cite{xnornet, xnornet++} have mostly revolved around convolutional neural networks. Recently, there has been some research on binarized Transformers. However, these studies have focused on machine translation or BERT pretraining, which is quite different from large language models.
For example, machine translation employs an encoder-decoder architecture, BERT pretraining utilizes a bidirectional encoder, and large language models use a unidirectional decoder.
Furthermore, large language models are typically scaled up to a much larger model size, while BERT and machine translation models do not undergo such extensive scaling.

To the best of our knowledge, this work is the first to investigate quantization-aware training for 1-bit large language models.
We propose \our{}, a 1-bit Transformer architecture for large language models, which aims to scale efficiently in terms of both memory and computation. \our{} employs low-precision binary weights and quantized activations, while maintaining high precision for the optimizer states and gradients during training. Our approach is designed to be scalable and stable, with the ability to handle large language models efficiently. The implementation of the \our{} architecture is quite simple, requiring only the replacement of linear projections (i.e., \emph{nn.Linear} in PyTorch) in the Transformer. Furthermore, it complements other acceleration methods for large language models, such as PagedAttention~\cite{pageattn}, FlashAttention~\cite{flashattn,flashattn2}, and speculative decoding~\cite{specdecoding}.

We evaluate \our{} on a range of language modeling benchmarks, comparing with state-of-the-art quantization methods and FP16 Transformers.
Experimental results demonstrate that \our{} achieves competitive performance in terms of both perplexity and downstream task accuracy. More importantly, \our{} significantly reduces memory footprint and energy consumption compared to the baselines.
Furthermore, we show that \our{} follows a scaling law similar to that of full-precision Transformers, indicating that it can be effectively scaled to even larger language models with potential benefits in terms of performance and efficiency.

\newpage

\section{\our{}}

\begin{figure}[t]
\centering
\includegraphics[width=\linewidth]{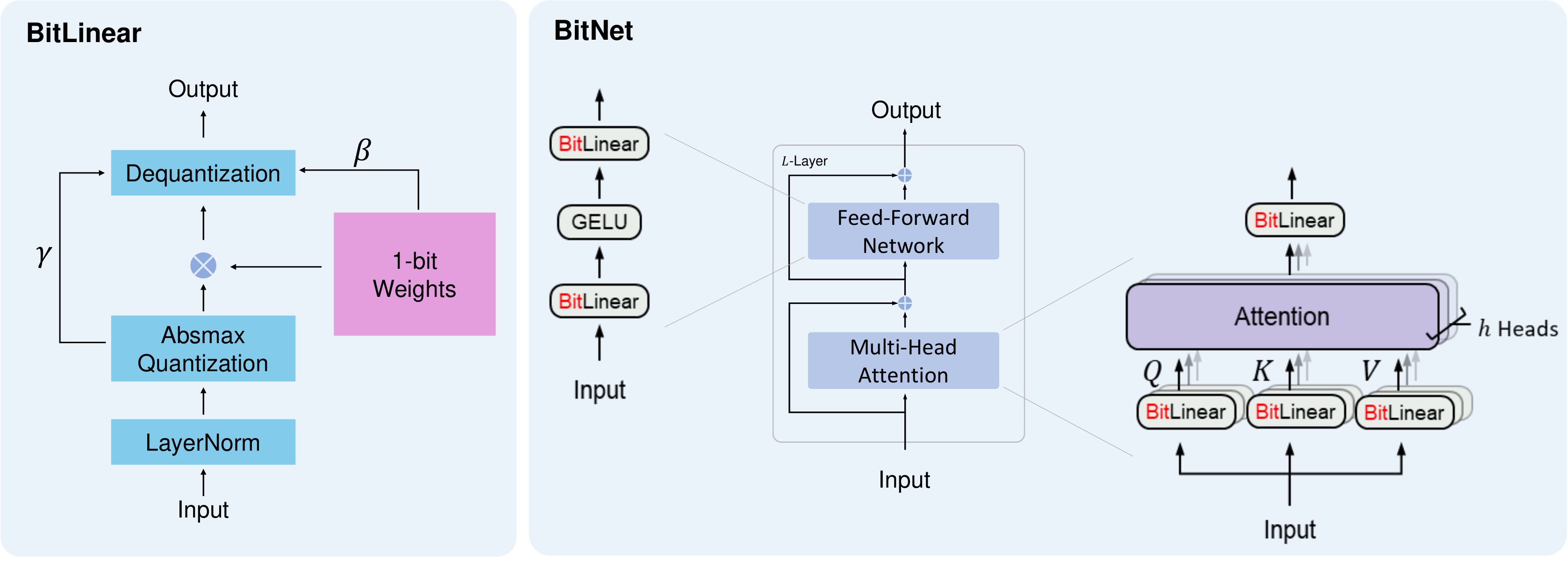}
\caption{(a) The computation flow of \texttt{BitLinear}. (b) The architecture of \our{}, consisting of the stacks of attentions and FFNs, where matrix multiplication is implemented as \texttt{BitLinear}.}
\label{fig:bitnet}
\end{figure}

As shown in Figure~\ref{fig:bitnet}, \our{} uses the same layout as Transformers, stacking blocks of self-attention and feed-forward networks.
Compared with vanilla Transformer, \our{} uses \texttt{BitLinear} (Eq.~\ref{eq_quant}) instead of conventional matrix multiplication, which employs binarized (i.e., 1-bit) model weights.
We leave the other components high-precision, e.g., 8-bit in our experiments.
We summarized the reasons as follows.
First, the residual connections and the layer normalization contribute negligible computation costs to large language models.
Second, the computation cost of QKV transformation is much smaller than the parametric projection as the model grows larger.
Third, we preserve the precision for the input/output embedding because the language models have to use high-precision probabilities to perform sampling.

\subsection{BitLinear}

We first binarize the weights to either $+1$ or $-1$ with the signum function. Following~\cite{bit}, we centralize the weights to be zero-mean before binarization to increase the capacity within a limited numerical range. A scaling factor $\beta$ is used after binarization to reduce the $l2$ error between the real-valued and the binarized weights. The binarization of a weight $W \in \mathcal{R}^{n \times m}$ can be formulated as:
\begin{equation}
    \widetilde{W} = \text{Sign}(W - \alpha),
    \label{eq_quant_weight}
\end{equation}

\begin{equation}
    \text{Sign}(W_{ij}) = \left\{
\begin{aligned}  
& +1, \quad && \text{if } W_{ij} > 0, \\  
& -1, \quad && \text{if } W_{ij} \leq 0, 
\end{aligned} 
\right.
\label{eq_sign}
\end{equation}
\begin{equation}
    \alpha = \frac{1}{nm}\sum_{ij} W_{ij}
\end{equation}

We further quantize the activations to $b$-bit precision. Following~\cite{llmint8}, we use absmax quantization, which scales activations into the range $[-Q_b, Q_b]$ ($Q_b=2^{b-1}$) by multiplying with $Q_b$ and dividing by the absolute maximum of the input matrix:

\begin{equation}
    \widetilde{x} = \mathrm{Quant}(x) = \mathrm{Clip}(x \times \frac{Q_b}{\gamma}, -Q_b+\epsilon, Q_b-\epsilon),
    \label{eq_quant_act}
\end{equation}

\begin{equation}
    \mathrm{Clip}(x, a, b) = \max(a, \min(b, x)), \quad \gamma = ||x||_{\infty},
    \label{eq_clip}
\end{equation}

where $\epsilon$ is a small floating-point number that prevents overflow when performing the clipping.

For the activations before the non-linear functions (e.g., ReLU), we scale them into the range $[0, Q_b]$ by subtracting the minimum of the inputs so that all values are non-negative:

\begin{equation}
    \widetilde{x} = \mathrm{Quant}(x) = \mathrm{Clip}((x-\eta) \times \frac{Q_b}{\gamma}, \epsilon, Q_b-\epsilon), \quad \eta = \min_{ij} x_{ij}.
\end{equation}

In this work, we quantize the activation to 8-bit and leave lower precision in future work. Moreover, the quantization is performed per tensor during training while per token during inference for both stability and efficiency.

With the above quantization equations, the matrix multiplication can be written as:

\begin{equation}
    y = \widetilde{W} \widetilde{x}
\end{equation}

We assume that the elements in $W$ and $x$ are mutually independent
and share the same distribution, and $W$ and $x$ are independent of each other. Then the variance of the output $y$ is estimated as:
\begin{align}
    \mathrm{Var}(y) &= n \mathrm{Var}(\widetilde{w} \widetilde{x}) \\
    &=n E[\widetilde{w}^2] E[\widetilde{x}^2] \\
    &=n \beta^2 E[\widetilde{x}^2] \approx E[\widetilde{x}^2]
\end{align}

For the full-precision computation, the variance of the output $\mathrm{Var}(y)$ is at the scale of $1$ with the standard initialization methods (e.g., Kaiming initialization or Xavier initialization), which has a great benefit to the training stability. To preserve the variance after quantization, we introduce a LayerNorm~\cite{layernorm} function before the activation quantization.
In this way, the variance of the output $y$ is then estimated as $\mathrm{Var}(y) \approx E[\mathrm{LN}(\widetilde{x})^2] = 1$, which has the same magnitude as the full-precision counterpart $\mathrm{Var}(y)$. In the context of Transformers, it has the exact implementation as \texttt{SubLN}~\cite{magneto}.
With \texttt{SubLN} and the quantization methods above, we have \texttt{BitLinear}, which is formulated as:

\begin{equation}
    y = \widetilde{W} \widetilde{x} = \widetilde{W}~\mathrm{Quant}(\mathrm{LN}(x)) \times \frac{\beta\gamma}{Q_b}\label{eq_quant}
\end{equation}

\begin{equation}
    \mathrm{LN}(x) = \frac{x-E(x)}{\sqrt{\mathrm{Var}(x)+\epsilon}}, \quad \beta = \frac{1}{nm}||W||_{1}
\end{equation}

Figure~\ref{fig:bitnet} provides an illustration of the computation flow of \texttt{BitLinear}. After the SubLN operation, the activations are quantized with the absmax function. The matrix multiplication is performed between the 1-bit weights and the quantized activations. The output activations are rescaled with $\{\beta, \gamma\}$ to dequantize them to the original precision.

\paragraph{Model parallelism with Group Quantization and Normalization}
One essential technique to scale up large language models is model parallelism~\cite{megatron}, which partitions the matrix multiplication on multiple devices. A prerequisite for the existing model parallelism approaches is that the tensors are independent along the partition dimension. However, all of the parameters $\alpha$, $\beta$, $\gamma$, and $\eta$ are calculated from the whole tensors, breaking the independent prerequisite. One solution is to introduce one \textit{all-reduce} operation for each parameter. However, even though the communication for each parameter is small, the amount of synchronization is growing as the model becomes deeper, which significantly slows the forward pass. The problem also exists in \texttt{SubLN}, where the mean and the variance should be estimated across the partition dimension.

To this end, we propose a simple approach that makes the model parallelism more efficient. 
We divide the weights and activations into groups and then independently estimate each group's parameters. This way, the parameters can be calculated locally without requiring additional communication. This approach, called Group Quantization, is formulated as follows:

For a weight matrix $W \in \mathcal{R}^{n \times m}$, we divide it into $G$ groups along the partition dimension, and each group has a size of $\frac{n}{G} \times m$. We then estimate the parameters for each group independently:

\begin{equation}
\alpha_g = \frac{G}{nm}\sum_{ij} W_{ij}^{(g)}, \quad \beta_g = \frac{G}{nm}||W^{(g)}||_1,
\end{equation}

where $W^{(g)}$ denotes the $g$-th group of the weight matrix. 
Similarly, for the activations, we can divide the input matrix $x \in \mathcal{R}^{n \times m}$ into $G$ groups and calculate the parameters for each group:

\begin{equation}
\gamma_g = ||x^{(g)}||_{\infty}, \quad \eta_g = \min_{ij} x_{ij}^{(g)}
\end{equation}

For LN, we can apply the group normalization technique~\cite{groupnorm} to compute the mean and variance for each group independently:

\begin{equation}
\mathrm{LN}(x^{(g)}) = \frac{x^{(g)}-E(x^{(g)})}{\sqrt{\mathrm{Var}(x^{(g)})+\epsilon}}
\end{equation}

In this way, we can efficiently implement model parallelism with Group Quantization and Normalization, which requires no additional communication and can scale to large language models.

\subsection{Model Training}

\paragraph{Straight-through estimator.}

To train our 1-bit model, we employ the straight-through estimator (STE)\cite{ste} to approximate the gradient during backpropagation. This method bypasses the non-differentiable functions, such as the Sign (Eq.~\ref{eq_sign}) and Clip (Eq.~\ref{eq_clip}) functions, during the backward pass. STE allows gradients to flow through the network without being affected by these non-differentiable functions, making it possible to train our quantized model. 

\paragraph{Mixed precision training.} 

While the weights and the activations are quantized to low precision, the gradients and the optimizer states are stored in high precision to ensure training stability and accuracy. Following the previous work~\cite{adambnn}, we maintain a latent weight in a high-precision format for the learnable parameters to accumulate the parameter updates. The latent weights are binarized on the fly during the forward pass and never used for the inference process.

\paragraph{Large learning rate.} 

One challenge for the optimization is that a small update on the latent weights often makes no difference in the 1-bit weights. This results in a biased gradient and update which are estimated based on the 1-bit weights. This problem is even worse at the beginning of the training, where the models are supposed to converge as fast as possible. To address this challenge, we explore various methods, concluding that increasing the learning rate is the simplest and best way to accelerate the optimization.
Our experiments show that \our{} benefits from a large learning rate in terms of convergence, while the FP16 Transformer diverges at the beginning of training with the same learning rate.
More details can be found in Section~\ref{experiment}.

\begin{table*}[t]
\setlength{\tabcolsep}{11pt}
\centering
\begin{tabular}{lccccccc}
\toprule
\multirow{2}{*}{Models} & \multirow{2}{*}{Size} & \multirow{2}{*}{WBits} & \multicolumn{2}{c}{7nm Energy (J)} & \multicolumn{2}{c}{45nm Energy (J)} \\
& & & MUL & ADD & MUL & ADD \\
\midrule
\multirow{2}{*}{Transformer} & \multirow{3}{*}{6.7B} & 32 & 4.41 & 1.28 & 12.46 & 3.03 \\
& & 16 & 1.14 & 0.54 & 3.70 & 1.35 \\
\our{} & & 1 & \bf 0.02 & \bf 0.04 & \bf 0.08 & \bf 0.13 \\
\midrule
\multirow{2}{*}{Transformer} & \multirow{3}{*}{13B} & 32 & 8.58 & 2.49 & 24.23 & 5.89 \\
& & 16 & 2.23 & 1.05 & 7.20 & 2.62 \\
\our{} & & 1 & \bf 0.04 & \bf 0.06 & \bf 0.12 & \bf 0.24 \\
\midrule
\multirow{2}{*}{Transformer} & \multirow{3}{*}{30B} & 32 & 20.09 & 5.83 & 56.73 & 13.80 \\
& & 16 & 5.21 & 2.45 & 16.87 & 6.13 \\
\our{} & & 1 & \bf 0.06 & \bf 0.14 & \bf 0.20 & \bf 0.53 \\
\bottomrule
\end{tabular}
\caption{Energy consumption of \our{} and Transformer varying different model size. Results are reported with 512 as input length.}
\label{tab:energy}
\end{table*}

\subsection{Computational Efficiency}
\label{sec:compute}

We estimate the computational efficiency of \our{} in terms of both arithmetic operations energy and memory footprint. We mainly focus on the calculation for the matrix multiplication, since it contributes the most to the cost of large language models.

\paragraph{Arithmetic operations energy.}

According to the energy model in~\cite{energycost,pokebnn}, the energy consumption for different arithmetic operations can be estimated as follows:

\begin{table*}[ht]
\setlength{\tabcolsep}{11pt}
\centering
\begin{tabular}{lccccccc}
\toprule
\multirow{2}{*}{Bits} & \multicolumn{2}{c}{ADD Energy $\hat{E}_{add}$ (pJ)} & \multicolumn{2}{c}{MUL Energy $\hat{E}_{mul}$ (pJ)}  \\
& 45nm & 7nm & 45nm & 7nm \\
\midrule
FP32 & 0.9 & 0.38 & 3.7 & 1.31 \\
FP16 & 0.4 & 0.16 & 1.1 & 0.34 \\
INT8 & 0.03 & 0.007 & 0.2 & 0.07 \\
\bottomrule
\end{tabular}
\caption{ADD and MUL energy consumption~\cite{energycost,pokebnn} for different bit representations at 45nm and 7nm process nodes.}
\label{tab:bit_energy}
\end{table*}

In vanilla Transformers, for matrix multiplication with dimensions $m \times n$ and $n \times p$, the energy consumption can be calculated as follows:

\begin{align}
E_{add} &= m \times (n-1) \times p \times \hat{E}_{add} \\
E_{mul} &= m \times n \times p \times \hat{E}_{mul}
\end{align}

For \our{}, the energy consumption of the matrix multiplication is dominated by the addition operations, as the weights are 1-bit.
The multiplication operations are only applied to scale the output with the scalars $\beta$ and $\frac{\gamma}{Q_b}$, so the energy consumption for multiplication can be computed as:
\begin{align}
E_{mul} &= (m \times p + m \times n) \times \hat{E}_{mul}
\end{align}
which is significantly smaller than that in Transformers.
The energy savings of W1A8 \our{} compared to a full-precision (32-32) and half-precision (16-16) Transformer are shown in Table~\ref{tab:energy}. As can be seen, \our{} provides significant energy savings, especially for the multiplication operations, which are the major component of the matrix multiplication energy consumption.

\section{Comparison with FP16 Transformers}
\label{experiment}

\subsection{Setup}
\label{sec:setup}

We train a series of autoregressive language models with \our{} of various scales, ranging from 125M to 30B. The models are trained on an English-language corpus, which consists of the Pile dataset, Common Crawl snapshots, RealNews, and CC-Stories datasets. We use the Sentencpiece tokenizer to preprocess data and the vocabulary size is 16K. Besides \our{}, we also train the Transformer baselines with the same datasets and settings for a fair comparison. More details can be found in the appendix.

\subsection{Inference-Optimal Scaling Law}

Neural language models have proven to scale predictably~\cite{scalinglawlm} with vanilla Transformer architecture. The loss scales as the power law with the amount of computation used for training. This allows us to determine the optimal allocation of a computation budget as well as predict the performance of large language models from smaller models. 

\begin{figure}[t]
    \centering
    \includegraphics[width=0.49\textwidth]{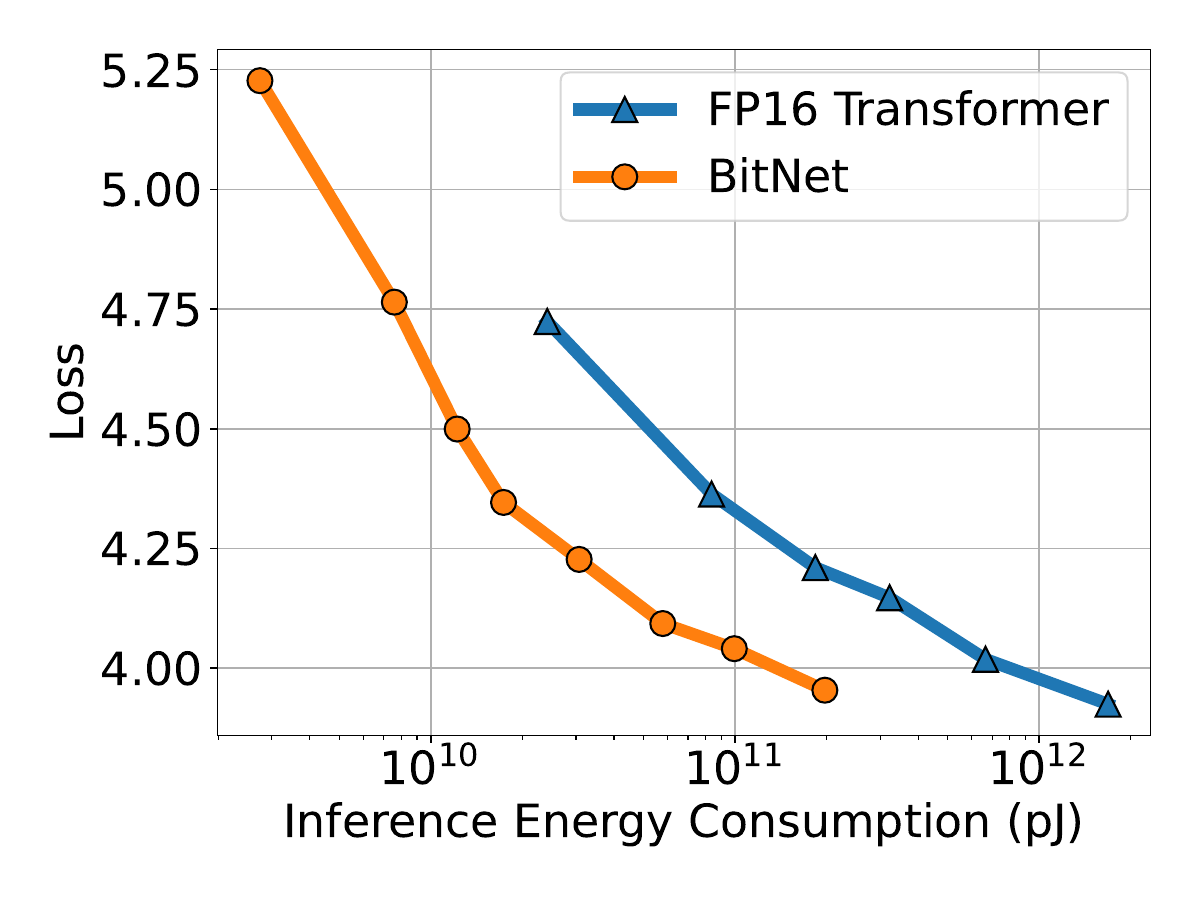}
    \includegraphics[width=0.49\textwidth]{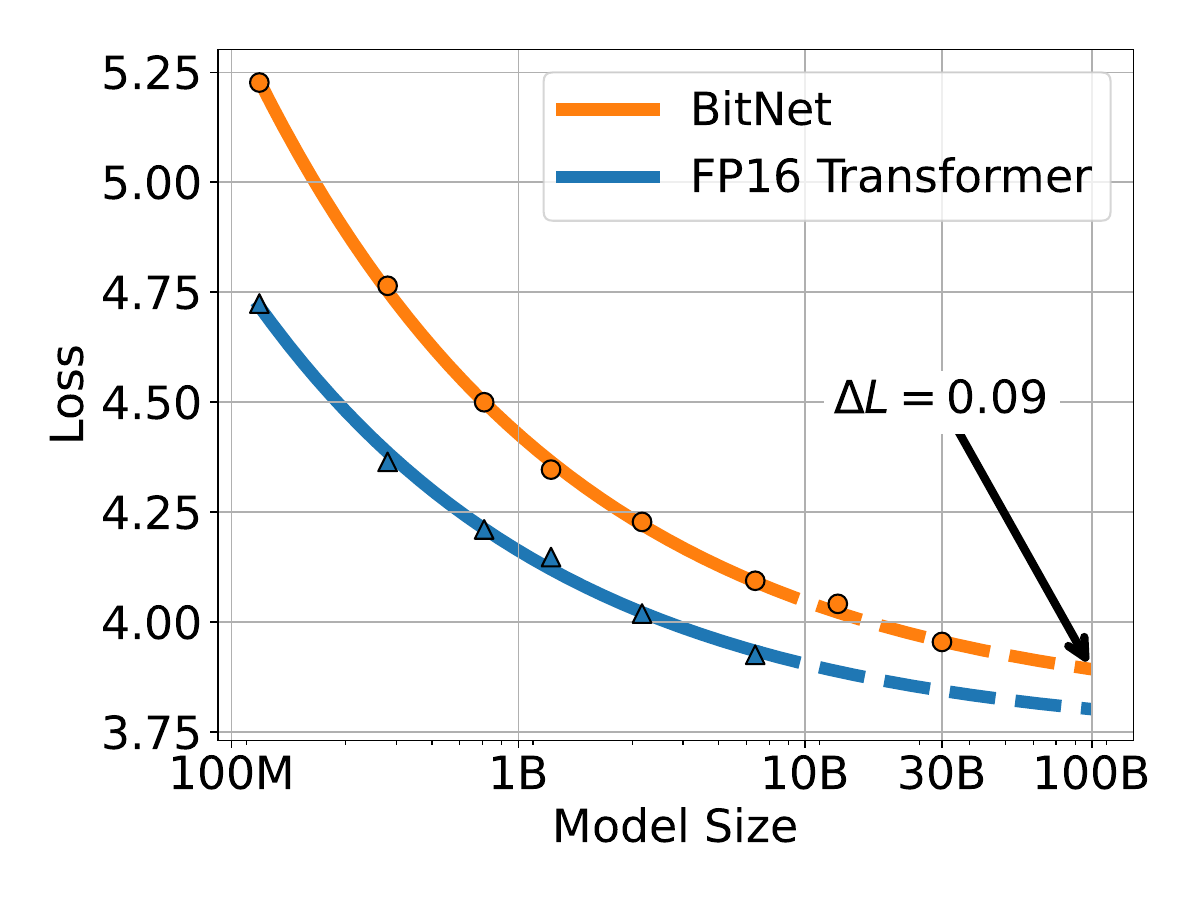}
    \caption{Scaling curves of \our{} and FP16 Transformers.}
    \label{fig:energy_vs_ppl}
\end{figure}

To study the scaling law of binarized Transformer, we start by plotting the scaling curve of both \our{} and the FP16 Transformer baseline against the parameter count. We fix the number of training tokens and vary the model sizes. Figure~\ref{fig:energy_vs_ppl} shows that the loss scaling of \our{} is similar to the FP16 Transformer, which follows a power-law. We then fit the scaling law with an irreducible loss term:
\begin{equation}
    L(N)=aN^b+c
\end{equation}
To evaluate whether the scaling law can accurately predict the loss, we choose the models from 125M to 6.7B to fit the parameters in the power-law and use the law to predict the loss of 13B and 30B. It shows that the fitted scaling law predicted \our{}'s loss with high accuracy. Besides, the gap between \our{} and FP16 Transformer becomes smaller as the model size grows.

While the power-law above measures the trend of the scaling of \our{}, it does not properly model the relationship between the loss and the actual compute. Previous work~\cite{scalinglawlm,scalinglawar,chinchilla} estimates the compute by calculating the FLOPs. However, it does not apply to 1-bit models whose cost is dominated by integer computation. Moreover, it mainly measures the training computation rather than the inference. To have a better understanding of the scaling efficiency of neural language models, we introduce Inference-Optimal Scaling Law. It predicts the loss against the energy consumption. We focus on the inference energy cost as it scales with the usage of the model, while the training cost is only once. We estimate the energy consumption as in Section~\ref{sec:compute}. Figure~\ref{fig:energy_vs_ppl} shows the scaling curve against the inference energy cost at 7nm process nodes. It proves that \our{} has much higher scaling efficiency. Given a fixed computation budget, \our{} achieves a significantly better loss. Meanwhile, the inference cost is much smaller to get the same performance as the FP16 models.

\subsection{Results on Downstream Tasks}

\begin{figure}[t]
    \centering
    \begin{subfigure}{.49\textwidth}
        \centering
        \includegraphics[width=\linewidth]{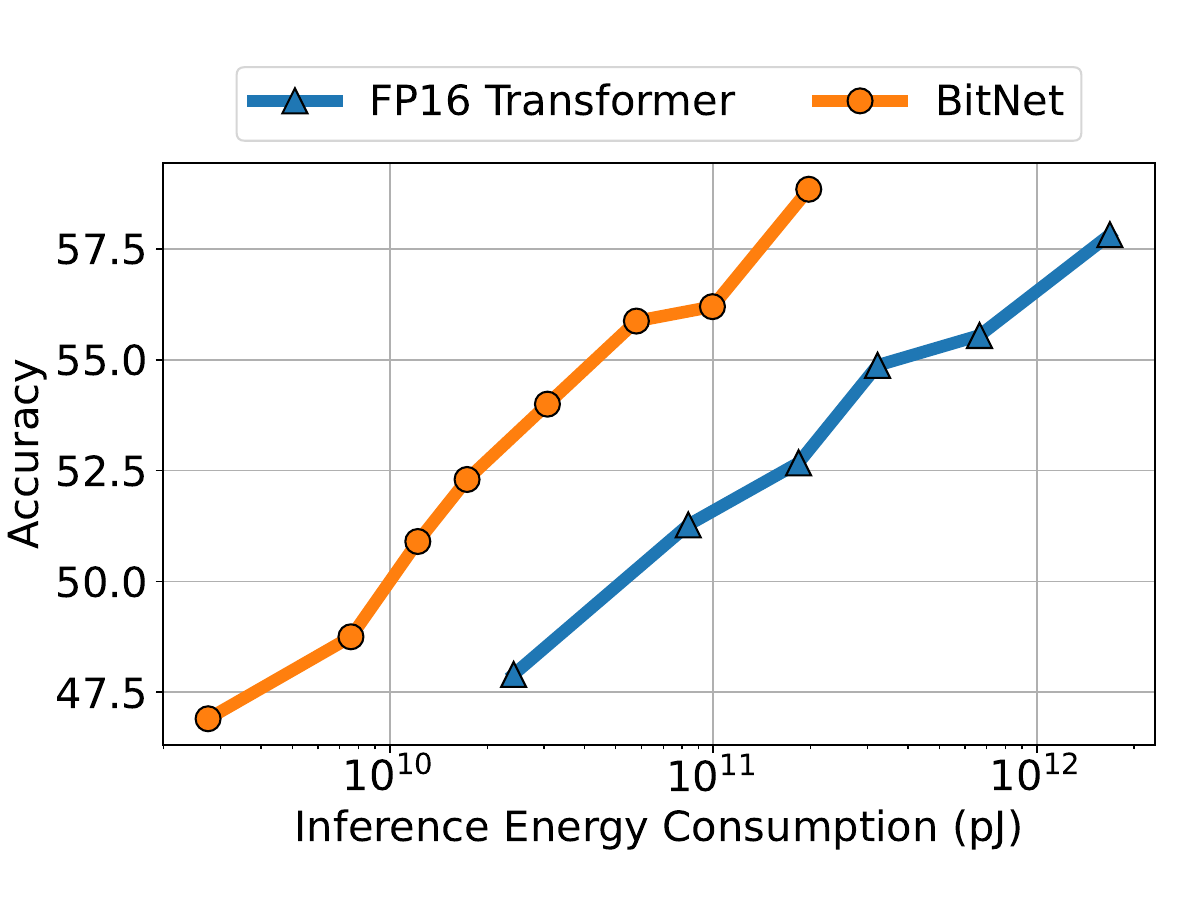}
        \caption{}
    \end{subfigure}
    \begin{subfigure}{.49\textwidth}
        \centering
        \includegraphics[width=\linewidth]{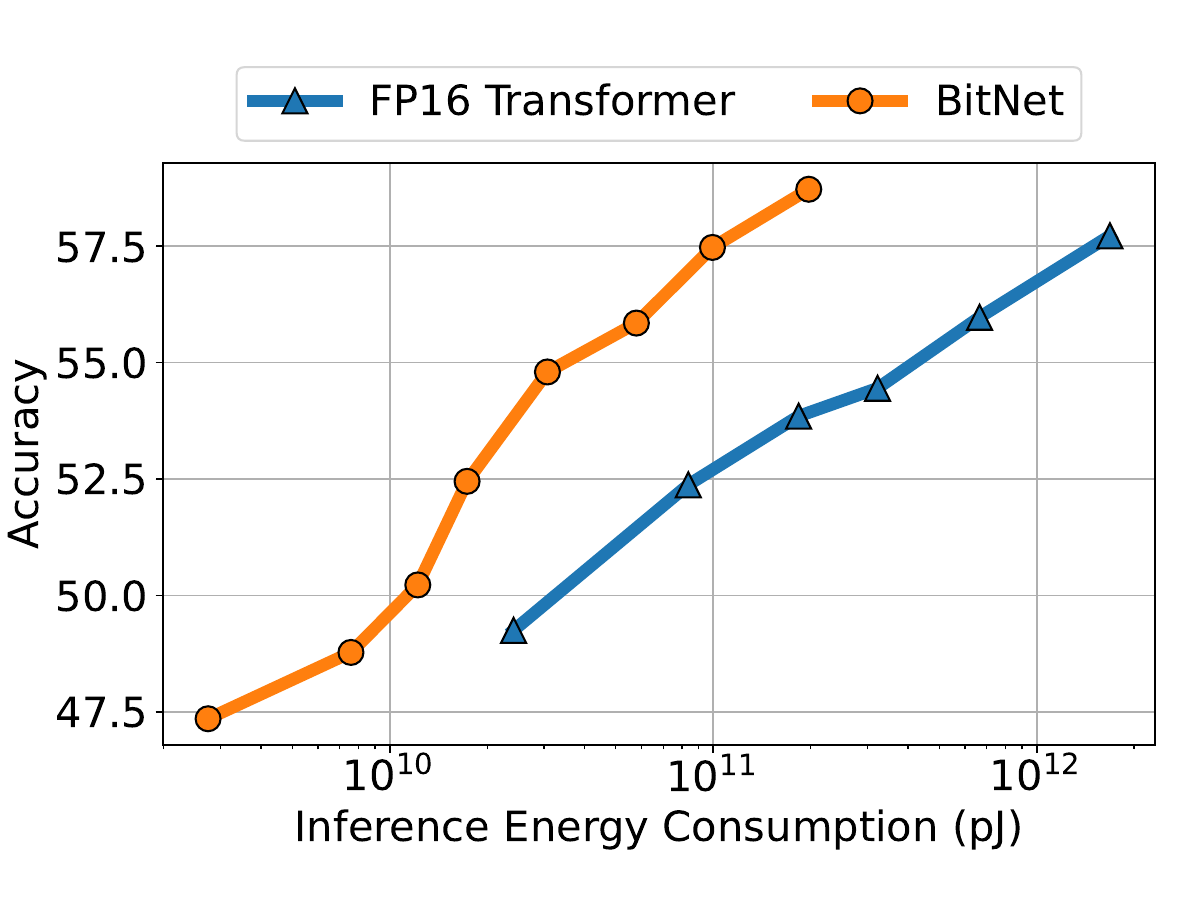}
        \caption{}
    \end{subfigure}
    \caption{Zero-shot (Left) and few-shot (Right) performance of \our{} and FP16 Transformer against the inference cost.}
    \label{fig:energy_vs_acc}
\end{figure}

In addition to the loss, we are also concerned about the capabilities with the scaling of \our{}. Compared with the loss, the capacity is more difficult to predict due to the emergent nature of neural language models. To evaluate the capabilities with the interpretable metrics, we test both the 0-shot and 4-shot results on four downstream tasks, including Hellaswag~\cite{hellaswag}, Winogrande~\cite{winoGrande}, Winograd~\cite{winograd}, and Storycloze~\cite{storycloze}. Figure~\ref{fig:energy_vs_acc} reports the average results of \our{} and FP16 Transformer with various scales. Similar to the loss scaling curve, the performance on the downstream tasks can scale as the computation budget grows. Besides, the scaling efficiency of capabilities is much higher than the FP16 Transformer baseline, in terms of both zero-shot and few-shot performance.

\subsection{Stability Test}

\begin{figure}[t]
    \centering
    \begin{subfigure}{0.495\textwidth}
        \centering
        \includegraphics[width=\linewidth]{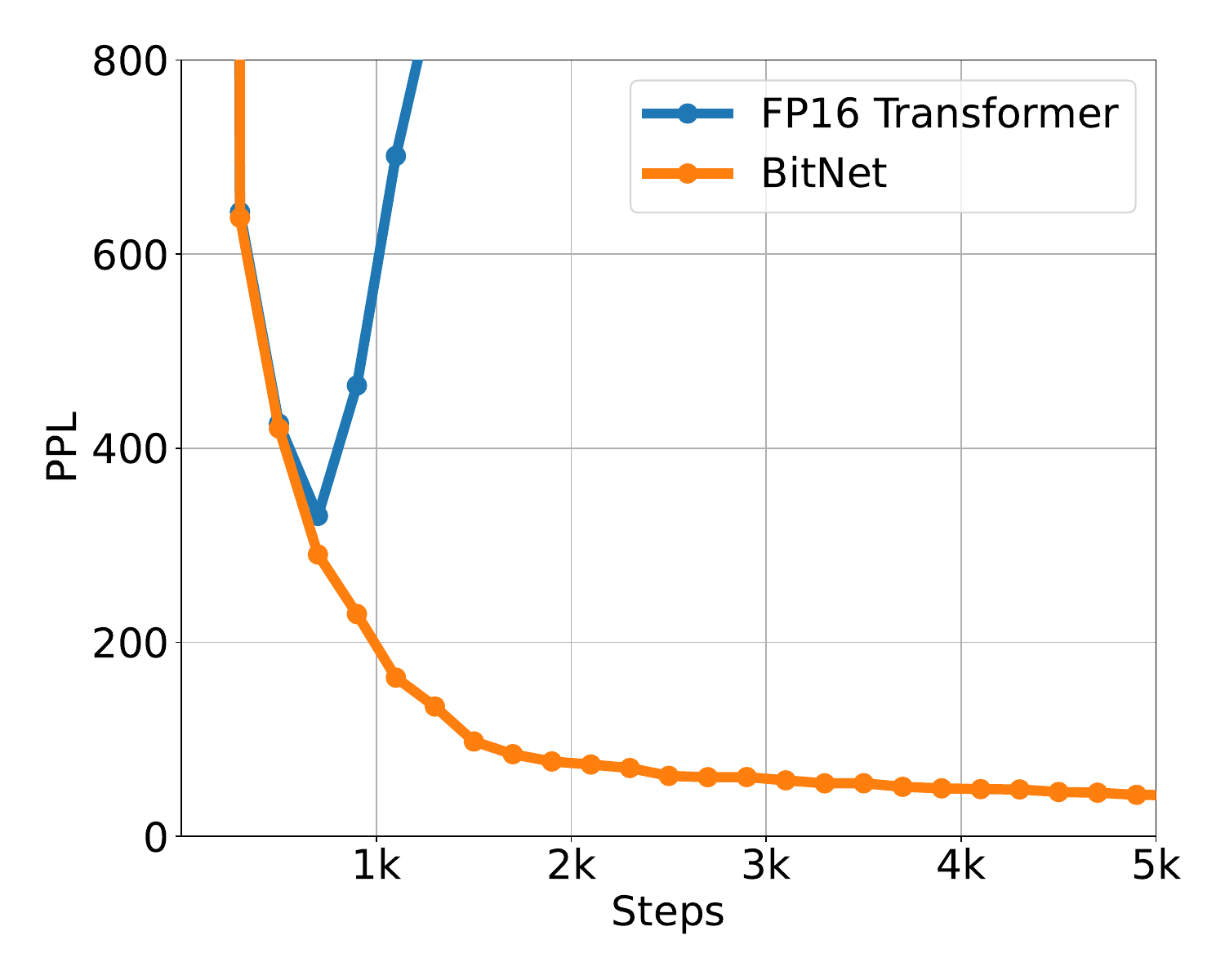}
        \caption{}
        \label{fig:stability}
    \end{subfigure}
    \begin{subfigure}{.495\textwidth}
        \centering
        \includegraphics[width=\linewidth]{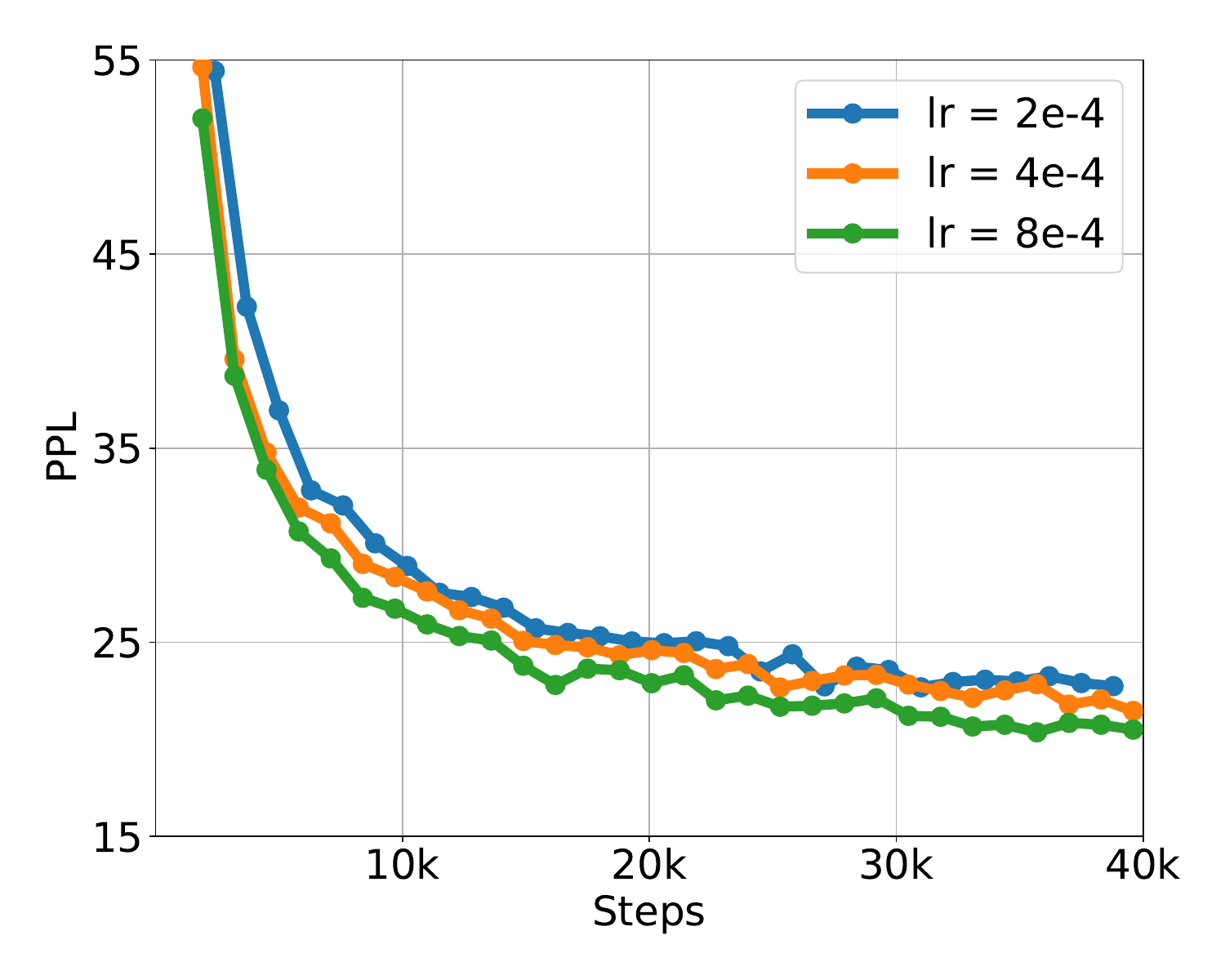}
        \caption{}
        \label{fig:high_lr}
    \end{subfigure}
    \caption{\our{} is more stable than FP16 Transformer with a same learning rate (Left). The training stability enables \our{} a larger learning rate, resulting in better convergence (Right).}
\end{figure}

The major challenge for training low-bit Transformers is the stability in optimization. Therefore, we perform stability tests for both \our{} and the FP16 baseline by training a series of models with varying peak learning rates. Figure~\ref{fig:stability} illustrates the results of the stability test. It shows that \our{} can converge with a large learning rate while FP16 Transformer can not, demonstrating better training stability of \our{}. This advantage in optimization enables the training with larger learning rates. Figure~\ref{fig:high_lr} shows that \our{} can benefit from the increase in learning rate, achieving better convergence in terms of PPL.

\section{Comparison with Post-training Quantization}

\subsection{Setup}

We train \our{} with the same setup as described in Section~\ref{sec:setup}. We compare \our{} with state-of-the-art quantization methods, including Absmax~\cite{llmint8}, SmoothQuant~\cite{smoothquant}, GPTQ~\cite{gptq}, and QuIP~\cite{quip}. These methods are post-training quantization over an FP16 Transformer model, which follows the same training setting and data as \our{}. Among them, Absmax and SmoothQuant quantize both the weights and the activations, while GPTQ and QuIP only reduce the precision of weights. We apply the methods to various quantization levels. For the weight-only quantization (i.e., GPTQ and QuIP), we experiment with W4A16 and W2A16. For weight-and-activation quantization (i.e., Absmax and SmoothQuant), we use them to quantize the FP16 Transformers to W8A8, W4A4, and W1A8. Our implementation of \our{} is binary weight 8-bit activation (W1A8), which has lower or equal bits than the baselines.

\subsection{Results}

Table \ref{tab:ptq} presents a detailed comparative analysis of the zero-shot performance of our proposed method, \our{}, against various baseline approaches on four benchmark datasets, namely Winogrande, Winograd, Storycloze, and Hellaswag. All models have the model sizes of 6.7B for a fair comparison. The methods are evaluated across several weight bit levels, spanning from 16 down to 1. Besides the zero-shot accuracy on the downstream tasks, the evaluation metrics include language model perplexity on the validation set, which provides a comprehensive understanding of each method's performance.

\begin{figure}[t]
    \centering
    \begin{subfigure}{\textwidth}
        \centering
        \includegraphics[width=0.85\linewidth]{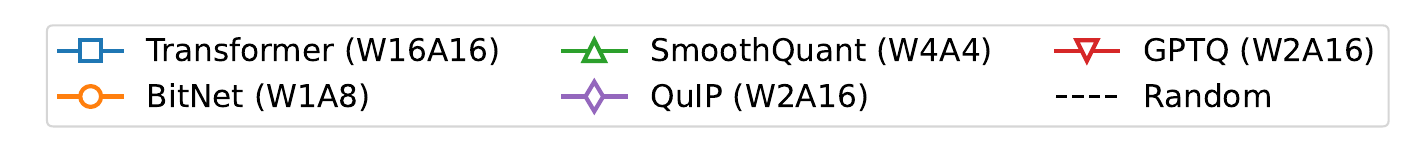}
    \end{subfigure}
    \begin{subfigure}{.495\textwidth}
        \centering
        \includegraphics[width=\linewidth]{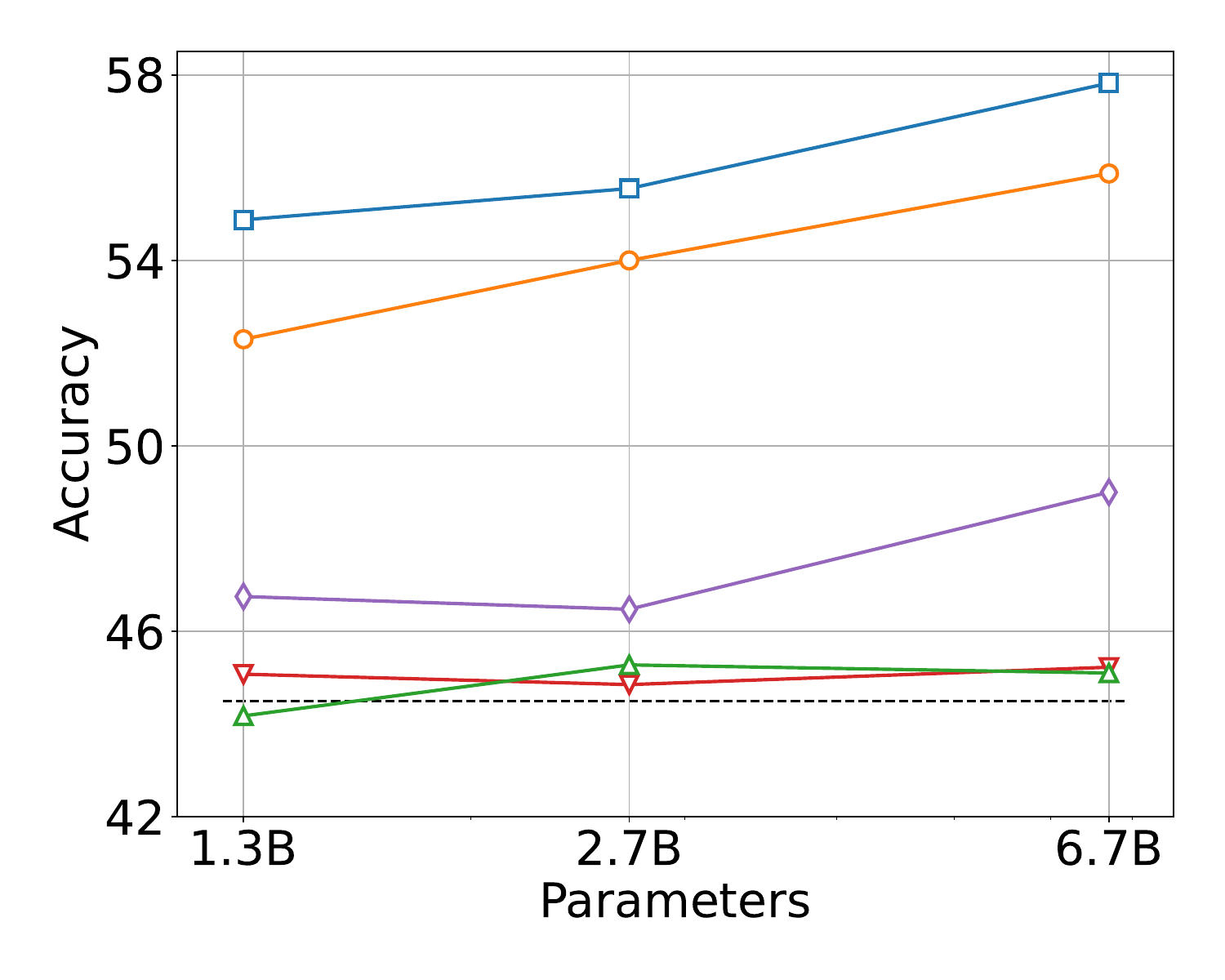}
        \caption{}
    \end{subfigure}
    \begin{subfigure}{.495\textwidth}
        \centering
        \includegraphics[width=\linewidth]{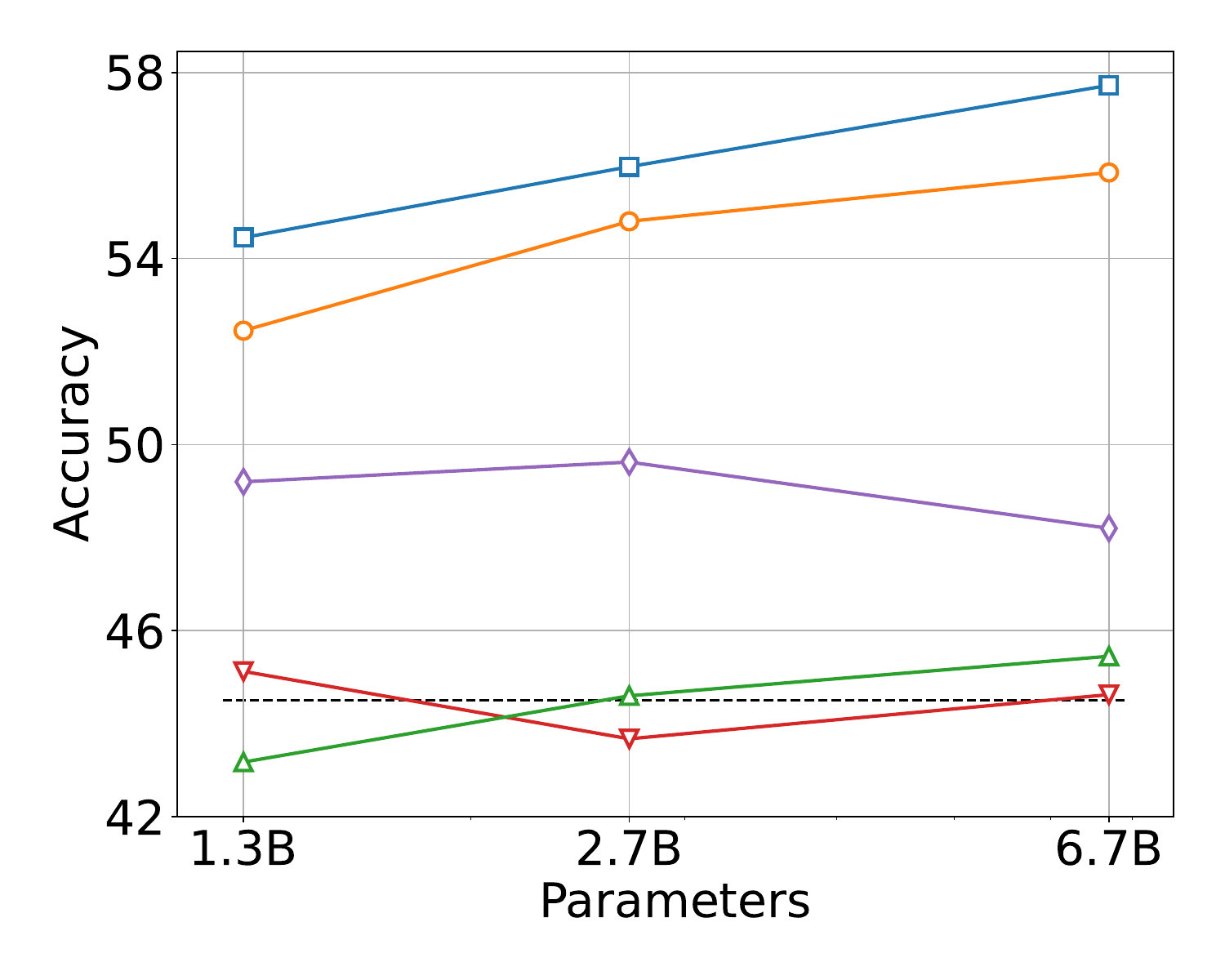}
        \caption{}
    \end{subfigure}
    \caption{Zero-shot (Left) and few-shot (Right) results for \our{} and the post-training quantization baselines on downstream tasks.}
    \label{fig:ptq}
\end{figure}

The results demonstrate the effectiveness of \our{} in achieving competitive performance levels compared to the baseline approaches, particularly for lower bit levels. The zero-shot scores of \our{} are comparable with the 8-bit models, while the inference cost is much lower. For the 4-bit models, the weight-only quantization methods outperform the weight-and-activation quantizers, mainly because the activation is more difficult to quantify. \our{}, as a 1-bit model, significantly achieves better results than both the weight-and-activation quantization methods and the weight-only methods. As for the lower-bit models, \our{} has consistently superior scores over all baselines. This proves the advantages of the quantization-aware training approaches over the post-training quantization methods. Figure~\ref{fig:ptq} summarizes both the zero-shot accuracy and few-shot accuracy of our method and the baselines while scaling up the model size from 1.3B to 6.7B. It proves that the advantage is consistent across different scales.

\begin{table*}[t]
\setlength{\tabcolsep}{8pt}
\centering
\begin{tabular}{clc|c|ccccc}
\toprule
\textbf{WBits} & \textbf{Methods} & \textbf{PTQ} & \textbf{PPL$\downarrow$} & \textbf{WG$\uparrow$} & \textbf{WGe$\uparrow$} & \textbf{HS$\uparrow$} & \textbf{SC$\uparrow$} & \textbf{Avg$\uparrow$}\\
\midrule
- & Random & \xmark & - & 50.0 & 50.0 & 25.0 & 50.0 & 43.8 \\
16 & Transformer & \xmark & 15.19 & 66.7 & 54.3 & 42.9 & 67.4 & 57.8 \\
\midrule
\multirow{2}{*}{8} & Absmax & \cmark & 21.43 & 60.4 & 52.0 & 38.3 & 62.7 & 53.4 \\
& SmoothQuant & \cmark & 15.67 & 65.3 & 53.1 & 40.9 & 67.6 & 56.7 \\
\midrule
\multirow{3}{*}{4} & GPTQ & \cmark & 16.05 & 57.2 & 51.2 & 39.9 & 63.4 & 52.9 \\
& Absmax & \cmark & 4.8e4 & 55.8 & 50.9 & 25.0 & 53.1 & 46.2 \\
& SmoothQuant & \cmark & 1.6e6 & 53.7 & 48.3 & 24.8 & 53.6 & 45.1 \\
\midrule
\multirow{2}{*}{2} & GPTQ & \cmark & 1032 & 51.6 & 50.1 & 25.8 & 53.4 & 45.2 \\
& QuIP & \cmark & 70.43 & 56.1 & 51.2 & 30.3 & 58.4 & 49.0 \\
\midrule
\multirow{2}{*}{1} & Absmax & \cmark & 3.5e23 & 49.8 & 50.0 & 24.8 & 53.6 & 44.6 \\
& SmoothQuant & \cmark & 3.3e21 & 50.5 & 49.5 & 24.6 & 53.1 & 44.4 \\
\midrule
1 & \bf \our{} & \xmark & 17.07 & 66.3 & 51.4 & 38.9 & 66.9 & 55.9 \\
\bottomrule
\end{tabular}
\caption{Zero-shot results for \our{} and the baselines (\texttt{PTQ}: Post-training quantization, \texttt{WGe}: Winogrande, \texttt{WG}: Winograd, \texttt{SC}: Storycloze, and \texttt{HS}: Hellaswag dataset).}
\label{tab:ptq}
\end{table*}

\section{Ablation Studies}

In Table \ref{tab:0_shot}, we present an ablation study of our compared with several alternative approaches. We ablate the effect of our choices in activation quantization approaches as well as the techniques to stabilize the model training. \our{} implement absmax to quantize the activation and use \texttt{SubLN} for training stability. One quantization alternative is the elastic function~\cite{bit}, which dynamically adjusts the scales with learnable parameters. In our experiments, we find that absmax has better performance than the elastic function. Besides, the absmax function leads to more stable training, which enables a larger learning rate for \our{}. We further compare \texttt{SubLN} with the Pre-LN and the BMT architecture~\cite{bmt}. Pre-LN is the default architecture for GPT pertaining, while BMT has proven to improve the stability of binarized models. Our experiments show that \texttt{SubLN} outperforms both Pre-LN and BMT. Therefore, we choose absmax and \texttt{SubLN} as the implementation in \our{}.

\begin{table*}[t]
    \setlength{\tabcolsep}{11pt}
    \centering
    \begin{tabular}{l|c|ccccc}
    \toprule
    \textbf{Methods}  & \textbf{PPL$\downarrow$} & \textbf{HS$\uparrow$} & \textbf{WGe$\uparrow$} & \textbf{WG$\uparrow$} & \textbf{SC$\uparrow$} & \textbf{Avg$\uparrow$}\\
    \midrule
    \multicolumn{6}{l}{\ \emph{Zero-Shot Learning}} \\
    \our{} & \bf 20.34 & 33.2 &  52.1 & 60.7 & 63.2 & \bf 52.3 \\
    \quad Elastic + Pre-LN & 24.05 & 29.6 & 52.9  & 56.8 & 61.3 & 50.2 \\
    \quad Absmax + Pre-LN  & 22.11 & 31.6 & 50.0 &  61.8 & 61.6 & 51.3 \\
    \quad Absmax + BMT & 22.98 & 31.2 & 52.1  & 60.4 & 62.7 & 51.6 \\
    
    \midrule
    \multicolumn{6}{l}{\ \emph{Few-Shot Learning}} \\
    \our{} &  \bf 20.34 & 33.5 & 50.4 & 62.1 & 63.8 & \bf 52.5\\
    \quad Elastic + Pre-LN & 24.05 & 29.9 & 51.7 & 57.5 & 61.1 & 50.1 \\
    \quad Absmax + Pre-LN & 22.11 & 31.4 & 51.9 & 63.9 & 61.6 & 52.2 \\
    \quad Absmax + BMT & 22.98 & 31.3 & 51.5 & 57.5 & 62.6 & 50.7 \\
    
    \bottomrule
    \end{tabular}
    \caption{Ablation of \our{} (\texttt{WGe}: Winogrande, \texttt{WG}: Winograd, \texttt{SC}: Storycloze, and \texttt{HS}: Hellaswag dataset). Elastic is an activation quantization method from \cite{bit}, while BMT is the architecture from \cite{bmt} to stabilize the training of low-bit models.}
    \label{tab:0_shot}
\end{table*}

\section{Conclusion and Future Work}
We present BitNet, a novel 1-bit Transformer architecture for large language models. Our approach is designed to be scalable and stable, with the ability to handle large language models efficiently. The experimental results demonstrate that BitNet achieves competitive performance in terms of both perplexity and downstream task performance, while significantly reducing memory footprint and energy consumption compared to the baselines. Moreover, BitNet follows a scaling law similar to that of full-precision Transformers, indicating that it can be effectively scaled to even larger language models with potential benefits in terms of performance and efficiency.
In the future, we would like to scale up \our{} in terms of
model size and training steps. We are also interested in applying \our{} in other architectures (e.g., RetNet~\cite{retnet}) for training large language models.


\bibliographystyle{alpha}
\bibliography{bitnet}


\appendix

\section{Hyperparameters}

\begin{table*}[ht]
    \setlength{\tabcolsep}{11pt}
    \centering
    \begin{tabular}{ccccc}
    \toprule
    \textbf{Params} & \textbf{\# Hidden} & \textbf{\# Layers} & \textbf{\# Heads} & \textbf{Learning Rate} \\
    \midrule
    125M & 768 & 12 & 12 & 2.4e-3 \\
    350M & 1024 & 24 & 16 & 1.2e-3 \\
    760M & 1536 & 24 & 16 & 1e-3 \\
    1.3B & 2048 & 24 & 32 & 8e-4 \\
    2.7B & 2560 & 32 & 32 & 6.4e-4 \\
    6.7B & 4096 & 32 & 32 & 4.8e-4 \\
    13B  & 5120 & 40 & 40 & 4e-4 \\
    30B  & 7168 & 48 & 56 & 4e-4 \\
    \bottomrule
    \end{tabular}
    \caption{Model configuration for \our{} in the scaling experiments.}
    \label{tbl:hyper:scaling}
\end{table*}

\begin{table*}[ht]
\centering
\begin{tabular}{l|c}
    \toprule
    \bf Hyperparameters & Value  \\
    \midrule
    Training updates & 40K \\
    Tokens per sample & 256K \\
    Adam $\beta$ & (0.9, 0.98) \\
    Learning rate schedule & Polynomial decay \\
    Warmup updates & 750 \\
    \midrule
    Gradient clipping & \ding{55} \\
    Dropout & \ding{55} \\
    Attention dropout & \ding{55} \\
    Weight decay & 0.01 \\
    \bottomrule
\end{tabular}
    \caption{
        Hyperparameters for \our{} and the FP16 Transformers in the scaling experiments. For 13B and 30B model, we set weight decay to 0.05 for training stability.
    }
\label{tbl:hyper:gpt}
\end{table*}

\begin{table*}[ht]
\centering
\begin{tabular}{l|c}
    \toprule
    \bf Hyperparameters &  Value \\
    \midrule
    Peak learning rate & 1e-3 \\
    Tokens per sample & 128K \\
    Adam $\beta$ & (0.9, 0.98) \\
    Learning rate schedule & Polynomial decay \\
    Warmup updates & 750 \\
    \midrule
    Gradient clipping & \ding{55} \\
    Dropout & \ding{55} \\
    Attention dropout & \ding{55} \\
    Weight decay & 0.01 \\
    \bottomrule
\end{tabular}
    \caption{
        Hyperparameters for the stability test of \our{} and FP16 Transformer. 
    }
\label{tbl:hyper:stabe_test}
\end{table*}

\begin{table*}[ht]
\centering
\begin{tabular}{l|cc}
    \toprule
    \bf Hyperparameters &  Elastic & Absmax\\
    \midrule
    Peak learning rate & 1e-4 & 8e-4 \\
    Training updates & \multicolumn{2}{c}{40K} \\
    Tokens per sample & \multicolumn{2}{c}{256K} \\
    Adam $\beta$ & \multicolumn{2}{c}{(0.9, 0.98)} \\
    Learning rate schedule & \multicolumn{2}{c}{Polynomial decay} \\
    Warmup updates & \multicolumn{2}{c}{750} \\
    \midrule
    Gradient clipping & \multicolumn{2}{c}{\ding{55}} \\
    Dropout & \multicolumn{2}{c}{\ding{55}} \\
    Attention dropout & \multicolumn{2}{c}{\ding{55}} \\
    Weight decay & \multicolumn{2}{c}{0.01} \\
    \bottomrule
\end{tabular}
    \caption{
        Hyperparameters for the ablations of \our{}. 
    }
\label{tbl:hyper:ablation}
\end{table*}

\end{document}